\pdfoutput=1

\documentclass[11pt]{article}
\usepackage[]{acl}

\usepackage{graphicx}
\usepackage{subcaption}

\usepackage{float}

\usepackage{algorithm}
\usepackage{algpseudocode}
\usepackage{tikz}

\usepackage{booktabs} 
\usepackage{tabularx} 
\usepackage{geometry} 
\usepackage{multirow}
\usepackage{times}
\usepackage{latexsym}
\usepackage{xspace}
\usepackage{amsmath}
\usepackage{amsfonts}
\usepackage{changepage} 
\usepackage[most]{tcolorbox}  
\usepackage{listings}
\usepackage{float}

\usepackage[T1]{fontenc}

\usepackage[utf8]{inputenc}

\usepackage{microtype}

\usepackage{inconsolata}

\usepackage{graphicx}

\newcommand{\ourtool}{\textbf{SemanticSpec}\xspace}

\title{Beyond Tokens: Semantic-Aware Speculative Decoding for Efficient Inference by Probing Internal States}

\author{Ximing Dong\textsuperscript{1}, Shaowei Wang\textsuperscript{2}\thanks{\;\;Corresponding author.}, Dayi Lin\textsuperscript{3}, Boyuan Chen\textsuperscript{1}, Ahmed E. Hassan\textsuperscript{3} \\
\textsuperscript{1}Centre for Software Excellence, Huawei, Canada \\
\textsuperscript{2} Department of Computer Science, University of Manitoba, Canada \\
\textsuperscript{3}School of Computing, Queen's University, Canada\\
\small
\texttt{\{ximing.dong, boyuan.chen1\}@huawei.com, shaowei.wang@umanitoba.ca,} \\
\small
\texttt{dayi.lin@queensu.ca, ahmed@cs.queensu.ca}
}

\begin{document}

%

\maketitle
\begin{abstract}
Large Language Models (LLMs) achieve strong performance across many tasks but suffer from high inference latency due to autoregressive decoding. The issue is exacerbated in Large Reasoning Models (LRMs), which generate lengthy chains of thought. While speculative decoding accelerates inference by drafting and verifying multiple tokens in parallel, existing methods operate at the token level and ignore semantic equivalence (i.e., different token sequences expressing the same meaning), leading to inefficient rejections. We propose \ourtool, a semantic-aware speculative decoding framework that verifies entire semantic sequences instead of tokens. SemanticSpec introduces a semantic probability estimation mechanism that probes the model’s internal hidden states to assess the likelihood of generating sequences with specific meanings. Experiments on four benchmarks show that \ourtool achieves up to 2.7× speedup on DeepSeekR1-32B and 2.1× on QwQ-32B, consistently outperforming token-level and sequence-level baselines in both efficiency and effectiveness. Our code is available at \url{https://github.com/Dongximing/semantic }.

\end{abstract}

\section{Introduction}\label{sec:intro}
\begin{figure*}
    \centering
    \includegraphics[width=0.9\linewidth]{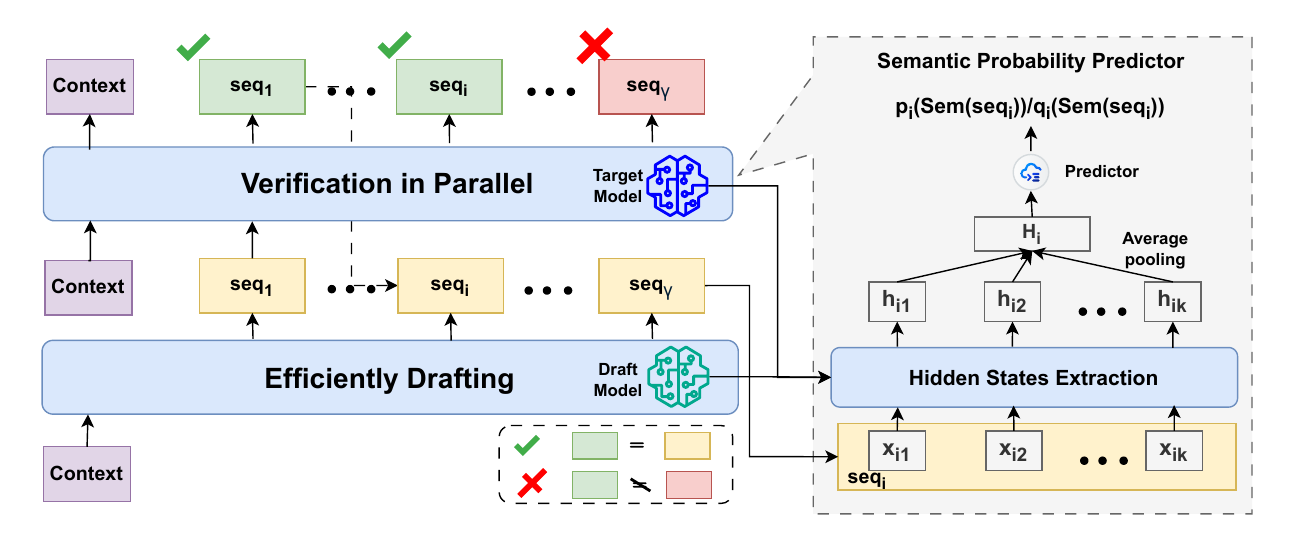}
    \caption{semantic-aware speculative decoding. Instead of drafting and verifying token-by-token, semantic-aware speculative decoding drafts and verifying sequence-by-sequence.}
    \label{fig:semantic speculative}
\end{figure*}

Large Language Models (LLMs) have achieved remarkable success across various tasks~\cite{openai2023gpt,bai2023qwen,touvron2023llama,liu2024deepseek,team2024gemma}. However, their deployment is constrained by high inference latency due to autoregressive decoding, where tokens are generated sequentially, and the latency grows with model size and output length. In particular, Large Reasoning Models (LRMs) such as OpenAI o1~\cite{jaech2024openai} and DeepSeek R1~\cite{guo2025deepseek} exacerbate this issue by producing lengthy chains of thought (CoT) for detailed intermediate reasoning before yielding the final output.


Speculative Decoding~\cite{spector2023accelerating,xia2024unlocking} has emerged as a promising approach to accelerate LLM inference. It drafts multiple future tokens and then verifies them in parallel using the target LLM. It only accepts those that meet its verification criterion to guarantee output quality. However, most existing methods operate at the token level and overlook semantic equivalence, where different token sequences that convey the same meaning (e.g., “The capital of France is Paris.” vs. “Paris is the capital of France.”)~\cite{kuhn2023semantic}. This oversight leads to unnecessary rejections of semantically valid drafts, and significantly reduces the efficiency of speculative decoding (as shown in Section~\ref{sec:rq1}). Our pilot study shows that in 56.5\% of the studied cases, the draft and target models agree on the reasoning intent, while traditional speculative decoding triggered a false rejection due to minor token-level differences~\ref{sec:falserejection}. Although recent work on Large Reasoning Models (LRMs) explores speculating reasoning steps instead of individual tokens, these approaches typically rely on LLM-as-a-judge evaluations, which are often biased and unreliable~\cite{gu2024survey,li2025evaluating,thakur2024judging}, and results in low performance.


To address this limitation, we propose semantic-aware speculative decoding (\ourtool), a novel framework that elevates speculative decoding from the token level to the sequence level and handles semantic-equivalence. \ourtool operates over entire semantic sequences, capturing higher-level meaning rather than surface token patterns. The key challenge lies in estimating how likely a model is to generate any sequence conveying a particular meaning, which cannot be directly accessed from token probabilities. To overcome this, \ourtool introduces a novel semantic probability (i.e., the likelihood of generating any sequence that expressing a particular semantic meaning) estimation mechanism by probing the internal hidden states of LLMs during the verification. Specifically, we develop a semantic probability predictor that learns to map the model’s internal representations when verifying a sentence, to the likelihood of producing the particular meaning expressed by the sentence. This predictor enables \ourtool to infer semantic-aware likelihoods beyond surface-level token distributions. 


To evaluate the effectiveness of \ourtool, we compare various token-level and sequence-level baselines (e.g., SpecReason~\cite{pan2025specreason}, Speculative Thinking~\cite{yang2025speculative}, and SpecSampling~\cite{yang2025speculative}) on four benchmarks (e.g., MATH-500 and GPQA-D) and two pairs of LLMs. Our empirical results show that \ourtool accelerates the target models, DeepSeekR1-32B and QwQ-32B, by an average of 2.7× and 2.1×, respectively, and consistently outperforms both studied sequence-level and token-level baselines in both effectiveness and efficiency. For instance, \ourtool further boosts throughput (i.e., Token Per Second) by 1.67X on the DeepSeekR1-32B and 2.66X on the QwQ-32B compared to the standard token-level speculative decoding.


\section{Background and Related Work}\label{sec:background}

\subsection{Speculative Decoding}


Speculative Decoding (SD) is a technique to accelerate LLM inference without changing the model's final output distribution in theory~\cite{speculativDecoding,xia2024unlocking}. Instead of querying the target LLM \(M_p\) for every token, a smaller and faster draft model \(M_q\) (e.g., a distilled or reduced-parameter version) speculatively generates a short token sequence; \(M_p\) then verifies this draft in one parallel forward pass. If the draft is correct, multiple tokens are accepted for the cost of a single forward pass.

The process for generating one block of tokens in basic speculative decoding~\cite{spector2023accelerating} is defined as follows:

\noindent\textbf{Step 1: Drafting.} Given a context sequence, the draft model (\(M_q\)) runs autoregressively for \(\gamma\) steps to propose a draft token sequence \(\tilde{x}_{1:\gamma} = (\tilde{x}_1, \tilde{x}_2, \dots, \tilde{x}_{\gamma})\). This is done by sampling from the draft model's distribution at each step:
$
\tilde{x}_i \sim M_q(\cdot \mid \text{context}, \tilde{x}_{1:i-1})
$


\noindent\textbf{Step 2: Parallel Verification.} The target model \(M_p\) takes \([\text{context}, \tilde{x}_{1:\gamma}]\) and performs a \textbf{single, parallel forward pass}, producing distributions \(p_{1:\gamma}\) for all draft positions:
$
p_{1:\gamma} = M_p(\cdot \mid \text{context}, \tilde{x}_{1:\gamma}).
$    

\noindent\textbf{Step 3: Acceptance \& Correction}
    The draft tokens are now verified sequentially. It accepts draft tokens until the first mismatch is found. The criterion for accepting the \(i\)-th draft token \(\tilde{x}_i\) is:
    If \(p_i(\tilde{x}_i) \geq q_i(\tilde{x}_i)\), accept \(\tilde{x}_i\). Otherwise, reject \(\tilde{x}_i\) and resample a corrected token \(x_i\) from an adjusted distribution ($norm(max(0, p(x) - q(x)))$). It discards all remained tokens and restarts the process from Step 1. 

\subsection{Related Work and Their Limitations}
\subsubsection{Token-level Speculative Decoding} 

The current token-level speculative decoding approaches fall into two families:

\noindent\textbf{Improving Drafting.}
Draft quality strongly affects SD efficiency. Existing work follows two paradigms: \emph{independent drafting} and \emph{self-drafting}. Independent drafting employs a smaller, separate model to propose candidate tokens for verification by the target LLM~\cite{xia2022speculative,leviathan2023fast,spector2023accelerating,sun2023spectr}. It often uses reduced-size models from the same family to accelerate larger ones without extra training. In contrast, self-drafting reuses the target model to generate drafts via internal sampling or parallel decoding~\cite{stern2018blockwise,cai2024medusa,yang2023predictive,zhang2023draft,hooper2025speed}. For instance, Medusa~\cite{cai2024medusa} attaches lightweight FFN heads for parallel token generation.
\textbf{Enhancing Verification.}
Each decoding step verifies drafted tokens in parallel to ensure consistency with the target model. Early methods used strict greedy verification that guaranteed lossless outputs~\cite{stern2019insertion,xia2022speculative} but frequently reject good drafts and limit speedup. Later work introduced more flexible criteria, e.g., top-k matching~\cite{xia2022speculative}, rollback strategies~\cite{kim2023speculative}, and token-tree verification~\cite{miao2023specinfer}, to improve acceptance and efficiency.

However, existing work either in improving drafting and verification remains token-level, overlooking semantic equivalence between drafts and target outputs, as discussed in Section~\ref{sec:intro}, which still potentially reject good candidates and leads to low acceptance rate.

\subsubsection{Sequence-level Speculative Decoding for Large Reasoning Models}

With the emergence of large reasoning models (LRMs), several approaches have been developed to speculate reasoning steps (i.e., sequences expressing reasoning steps), which consider the semantic equivalence rather than individual tokens~\cite{pan2025specreason,yang2025speculative,wang2025efficient,liao2025reward}. For example, Pan et al. proposed SpecReason~\cite{pan2025specreason}, which employs a draft model to perform speculative reasoning (e.g., reasoning steps) and prompts the target model to score the reasoning quality on a 1–10 scale for verification. Similarly, Yang et al. introduced Speculative Thinking~\cite{yang2025speculative}, which leverages the occurrence of reasoning-supportive tokens, such as ``wait'' or ``alternatively'', as cues to delegate the remaining reasoning process (e.g., reasoning steps) to a larger mentor model.
However, existing approaches mainly rely on LLM-as-a-judge evaluations or hard-coded lexical signals (e.g., “wait,” “alternatively”) for verification, both of which are prone to inaccuracies, as prior studies have shown that LLM-as-a-judge can be biased and unreliable~\cite{gu2024survey,li2025evaluating,thakur2024judging}. In contrast, our method leverages the internal states of the target model to measure the alignment between the draft and target reasoning processes.

\section{Methodology}\label{sec:method}
\newcommand{\commentline}[1]{%
  \Statex \textcolor{blue}{#1}%
}

We introduce Semantic-Aware Speculative Decoding (\ourtool), a new paradigm that re-conceptualizes speculative decoding at the level of semantic meaning, rather than individual tokens. While traditional methods sequentially draft and verify token-by-token, \ourtool drafts and verifies sequences, meanwhile handling semantic-equivalence robustly. Unlike token-level approaches, where token probabilities are directly accessed from LLMs, determining the probability of generating any sequence that conveys a specific semantic meaning (i.e., called \textbf{\textit{semantic probability}}) is challenging. Therefore, to effectively verify whether the drafted sequences align with the target model's true intent, \ourtool must estimate the semantic probability. We overcome this by designing a novel semantic probability predictor that infers these probabilities by probing the LLM's internal representations (hidden states). To train the predictor, we prompt LLMs with diverse inputs and generate multiple outputs and their hidden states. We then cluster outputs by semantic meaning and estimate their semantic probability.

\subsection{Sematic-Aware Speculative Decoding}
\begin{algorithm}[t]
\small

\caption{Semantic-Aware Speculative Decoding}
\label{alg:Semantic-speculative-decoding}
\begin{algorithmic}[1]
\Require Target model $\mathcal{M}_p$, draft model $\mathcal{M}_q$, input sequences $s_{1:t}$, number of sequences generated by draft model $\gamma$, probability predictor for draft and target model $Predictor_q$ and $Predictor_p$, max new tokens $T$  
\Ensure Generated sequence $s_{1:T}$

\commentline{\#Initialize position counter and Maximum number of draft sequences}
\State $n \leftarrow t$ 
\While{$len(s_{1:n}) < T$}
    \State $(\tilde{s}_{n+1},h^q_{n+1}), \ldots, (\tilde{s}_{n+\gamma},h^q_{n+\gamma})$ $\leftarrow$ $\text{DRAFT}(s_{\leq n}, \mathcal{M}_q)$

    \commentline{\#Extract hidden states in parallel from target model}
    \State $h^p_i \leftarrow \mathcal{M}_p(\tilde{s}_{n+i}|s_{\leq n+i-1})$ for $i=1,\ldots,\gamma $

    \commentline{\#Predict the semantic probability of  draft model and target model}

    \State $q_i(Sem(\tilde{s}_{n+i})|s_{\leq n+i-1}) \leftarrow Predictor_q(h^q_{n+i})$  for $i=1,\ldots,\gamma$
    \State $p_i(Sem(\tilde{s}_{n+i})|s_{\leq n+i-1}) \leftarrow Predictor_p(h^p_{n+i})$  for $i=1,\ldots,\gamma$

    \commentline{\#Sequence by sequence verification}
    
    \For{$i = 1$ to $\gamma$}
        \If{With probability $\min\left(1, \frac{p_i(Sem(\tilde{s}_{n+i})|s_{\leq n+i-1})}{q_i(Sem(\tilde{s}_{n+i})|s_{\leq n+i-1})}\right)$}
            \State Accept $\tilde{s}_{n+i}$
        \Else
            \State $s_{n+i} \leftarrow \mathcal{M}p(s|s_{1:n+i-1})$
            \State \textbf{break}
        \EndIf
    \If{$EOS$ in $s_{n+i}$}
    \State Return FirstTokens($s_{1:n+i}$,min(index($EOS$),T)
    \EndIf
    \EndFor
    \State $n \leftarrow n + i$

\EndWhile

\State Return FirstTokens($s_{1:n}$,T)

\end{algorithmic}
\end{algorithm}

Algorithm~\ref{alg:Semantic-speculative-decoding} outlines the process of Semantic-Aware Speculative Decoding implemented by our approach (\ourtool). The algorithm consists of three main stages: (1) \textbf{sequence drafting}, (2) \textbf{semantic-aware sequence verification}, and (3) \textbf{acceptance decision}. 

In sequence drafting, the draft model \(M_q\) speculatively 
generates up to \(\gamma\) candidate sequences 
\(\tilde{s}_{n+1}, \dots, \tilde{s}_{n+\gamma}\) based on the current context \(s_{1:n}\). 
Along with these drafted sequences, the hidden representations 
\(\{h^q_{n+1}, \dots, h^q_{n+\gamma}\}\) of the draft model are extracted to capture 
the internal state associated with each generated sequence (line 4). 
In this study, we use \texttt{\textbackslash n\textbackslash n} as split to separate sequences.

After drafting, the target model \(M_p\) verifies the drafted sequences by processing them 
in parallel to obtain their corresponding hidden states 
\(\{h^p_{n+1}, \dots, h^p_{n+\gamma}\}\) (line 5). Later, we use semantic probability predictors—\(Predictor_q\) for the draft model and \(Predictor_p\) for the target model to estimate the likelihood that each model would produce the semantic meaning expressed by the given sequence \(\tilde{s}_{n+i}\) un
der the same context (lines 6-7). The resulting probabilities \(q_i(Sem(\tilde{s}_{n+i})|s_{\leq n+i-1})\) and \(p_i(Sem(\tilde{s}_{n+i})|s_{\leq n+i-1})\) 
represent the semantic-aware confidence of each model in producing a particular semantic meaning, 
which forms the basis for subsequent acceptance decisions.
 
Lastly, for each drafted sequence, \ourtool\ determines whether to accept it based on the agreement 
between the draft and target models. Specifically, a drafted sequence 
\(\tilde{s}_{n+i}\) is accepted with probability 
\(\min\left(1, \frac{p_i(Sem(\tilde{s}_{n+i})|s_{\leq n+i-1})}{q_i(Sem(\tilde{s}_{n+i})|s_{\leq n+i-1})}\right)\) (lines 9-11).
Otherwise, the algorithm reverts to the target model \(M_p\) to regenerate the next sequence 
from the last accepted context, ensuring correctness while maintaining efficiency (lines 12-16). 
This verification–acceptance loop continues until the max token number or $EOS$ is met.

\subsection{Predicting Semantic Probability by Probing Hidden States of LLMs}

As motivated in Section~\ref{sec:background}, one challenge is that an LLM may produce different sequences to express the same meaning (i.e., semantic equivalence)~\cite{kuhn2023semantic}. Directly estimating the likelihood of a sequence of tokens (i.e., lexical representation) is problematic. 
In almost all applications, we care about meaning rather than lexical representation. Therefore, to overcome this challenge, we convert the problem of estimating the likelihood of generating a specific sequence to estimating the probability of generating a semantic meaning expressed by the sequence, i.e.,  semantic probability. Formally, given a context $C$ and a target LLM $\mathcal{M}$, our goal is to construct a predictor to estimate the likelihood of generating the semantic meaning expressed in the sequence $s$ generated by $\mathcal{M}$ 
$p(Sem(s)|[C,s],\mathcal{M})$ 
, where $Sem(s)$ is the semantic meaning expressed by $s$. Therefore, the semantic probability is an aggregated probability can be calculated: 
\[p(Sem(s)|[C,s],\mathcal{M}) = \sum_{s_i \in \text{Sequences}(S)}{p(s_i|C,\mathcal{M})} \] 
,where $\text{Sequences}(S)$ is the set of all possible token sequences that express the semantic meaning $Sem(s)$.

\begin{figure}[t]
    \centering

    \begin{subfigure}{0.5\textwidth}
        \includegraphics[width=\linewidth]{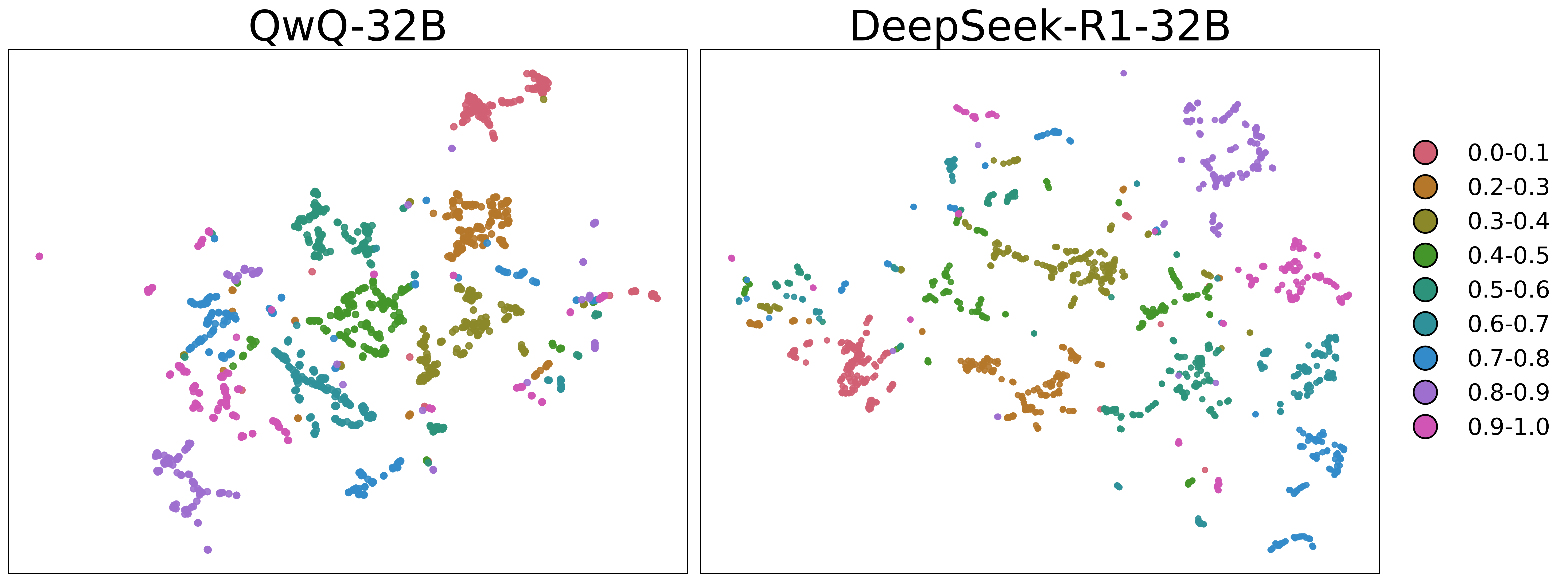}
   
    \end{subfigure}

    \caption{The distribution of hidden states with a various range of semantic probability in two-dimension. For each 0.1 probability interval, we randomly sample hidden state vectors, then project them into two dimensions using UMAP.}
    \label{fig:invalideProp}
\end{figure}

Intuitively, the semantic probability for a given context could be empirically estimated through a sampling approach: generating a large number of model outputs, clustering those outputs based on their semantic equivalence, and then using the relative frequency of each cluster as its probability estimate. However, this sampling during inference is cost inefficiency and violates the objective of speculative decoding. Prior studies demonstrate that the internal states of an LLM can indicate the confidence/uncertainty of output generated the LLM~\cite{kuhn2023semantic,azaria-mitchell-2023-internal,kossen2024semanticentropyprobesrobust}. To assess whether the internal states of LLMs can effectively reflect semantic probabilities in our setting, we examine relationship between internal states and semantic probability. As shown in Figure~\ref{fig:invalideProp}, our results reveal a pronounced clustering effect, where hidden states are organized into distinct regions according to their semantic probability levels. In other word, hidden states can effectively indicate the probability.

Therefore, we develop a semantic probability predictor that, by probing the internal states of the LLM without the need for sampling (see more details in Section~\ref{sec:predictor}). More importantly, the predictor can be trained offline and enables efficient semantic-aware verification during decoding. Below, we elaborate on the predictor and its construction process in detail.

\subsubsection{Offline Semantic Probability Predictor Training}\label{sec:predictor}

To train the semantic probability predictor, we first need to construct training data $
\mathcal{D} = \{ (H_{ij}, p(Sem(s_i)|[C_j,s_i],\mathcal{M})) \}_{i=1:N,j=1:M}$, which is a set of pair of semantic probability $p(Sem(s_i)|[C_j,s_i],\mathcal{M}))$ and its corresponding hidden states, where $s_i = (x_{i1},x_{i2}, \dots, x_{ij}, \dots, x_{ik})$ is a sequence that is verified by the LLM $\mathcal{M}$ given the input context $C_j$. The hidden states $H_{ij}$ is the aggregated representation derived from when the sequence $s_i$ goes through the model. More specifically, we collect the hidden states of the model when each token goes through the model.  Then to aggregate the hidden states from each token and obtain a fixed-length representation for a sequence, we apply average pooling over the token-level hidden states for aggregation. In this study, we use the hidden states from all layers of $\mathcal{M}$, which has been demonstrated superior than that from the last layer (see more details in Section~\ref{sec:rq2}). 

To construct the training data, we prompt the LLM $\mathcal{M}$ with various contexts (e.g., original query + multiple reasoning steps). For a specific context input $C_j$, we perform the following steps to estimate their semantic probability by following prior study~\cite{kuhn2023semantic}:

\noindent\textbf{Step 1. Generation:} Generating $N$ output sequences \{$s_1$, $\dots$, $s_N$\} using the large language model given a context $C_j$. 

\noindent\textbf{Step 2. Clustering:} Clustering sequences that share the same meaning using the \textit{bi-directional entailment} algorithm~\cite{kuhn2023semantic}. Two sequences, \(s\) and \(s'\), are considered semantically equivalent if and only if each entails the other (i.e., they logically imply one another). 
For example, ``The capital of France is Paris.'' entails ``Paris is the capital of France.'' because both express the same proposition. In our implementation, we employ the \texttt{DeBERTa-large} model to perform entailment detection.  For each pair of sampled sequences \(s\) and \(s'\), we check whether the concatenation of the context and \(s\) can be inferred from the concatenation of the context and \(s'\), and vice versa. Specifically, we concatenate each question–answer pair, then combine both sequences with a special separator token.  The DeBERTa model classifies this composite input into one of three categories:  \textit{entailment}, \textit{neutral}, or \textit{contradiction}. Two sequences are deemed \textbf{semantically equivalent} if and only if both directional inferences result in \textit{entailment}. We cluster a pair of two sequences together if they are semantically equivalent. 

\noindent\textbf{Step 3. Probability estimation:} After clustering, we obtain a set of semantic clusters 
\(\{C_1, C_2, \dots, C_K\}\), 
where each cluster contains sequences that convey the same meaning. We then estimate the probability of generating a specific semantic meaning by counting the size of the cluster. More specifically, 
for any sequence $s_i \in C_k$, its semantic probability can be calculated: 
\[p(Sem(s_i)|[C_i,s_i],\mathcal{M}) = \frac{|C_k|}{N}\]

Suppose there are $M$ contexts and for each context, we generate $N$ outputs, then the size of the training data is $MN$. Once we construct the training data, we train a regression model, which takes the hidden states as the input and output their corresponding semantic probability. We deploy a multilayer perceptron with three layers as our regression model. Note that we train two predictors, one for the target model and one for the draft model.

\subsubsection{Online Semantic Verification}
During online inference, the trained \textit{semantic probability predictor} verifies whether a drafted sequence aligns with the target model’s intent. For each drafted sequence, we extract its hidden representation and predict the semantic probability using the predictor for draft model. The same sequence is processed by the target model to obtain corresponding hidden states, and the probability is predicted. The acceptance of a drafted sequence is determined as shown in Algorithm~\ref{alg:Semantic-speculative-decoding}, where a higher ratio indicates stronger semantic agreement. This semantic verification captures semantic consistency between models, allowing speculative decoding to remain robust to lexical variations while improving efficiency.

\begin{figure*}[h]
    \centering
    \includegraphics[width=0.95\linewidth]{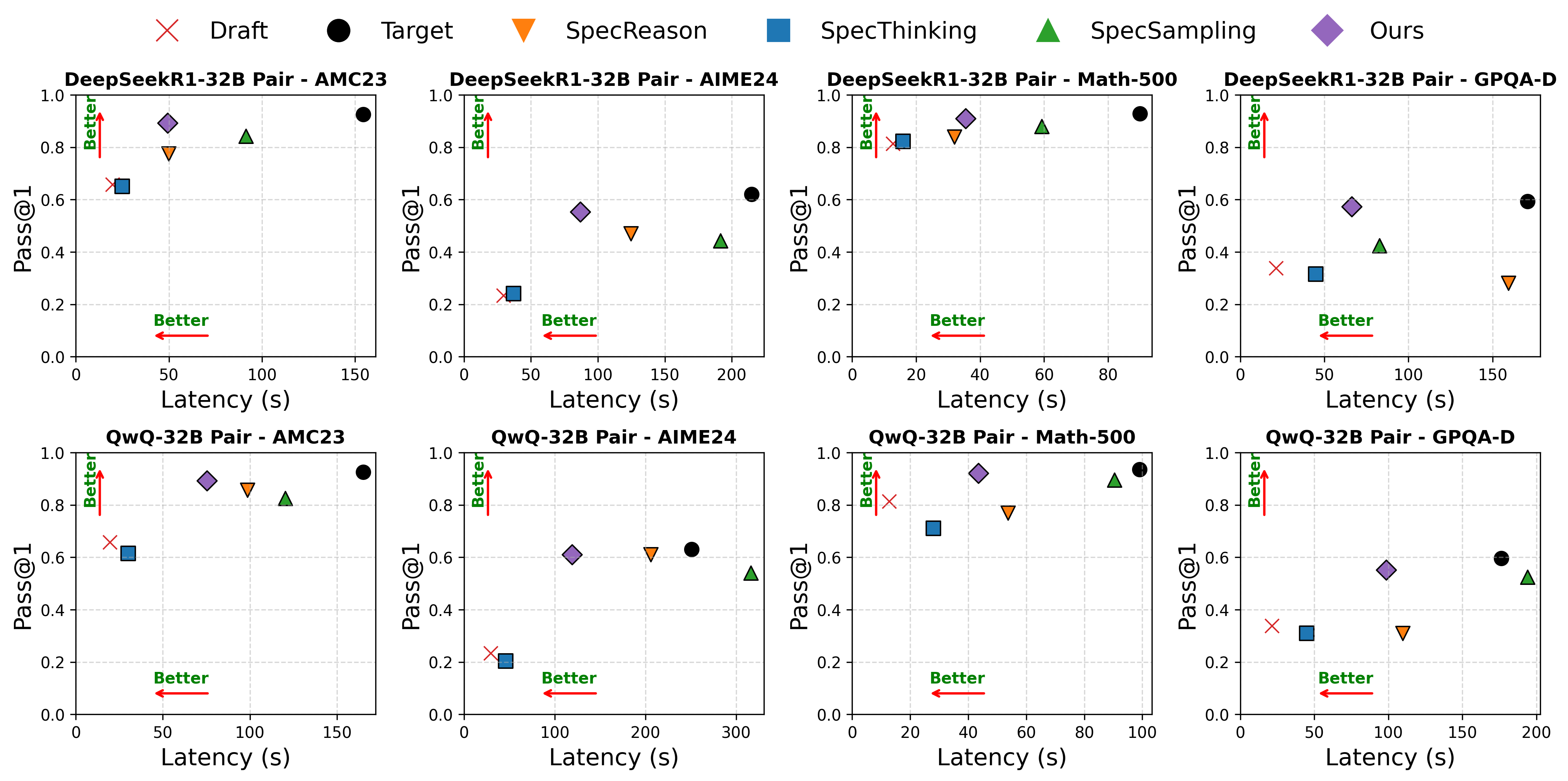}
    \caption{Comparison of pass@1 and latency across different speculative inference schemes and model pairs.
    Each point represents a decoding approach.
    The top row shows results for the target model DeepSeekR1-32B, and the bottom row for QwQ-32B.
    }
    \label{fig:latency_acc_pairs}
\end{figure*}

\section{Experimental Settings}\label{lab:experimentalsetting}
 
\begin{table*}[ht]
\centering
\resizebox{2.05\columnwidth}{!}{
\renewcommand{\arraystretch}{1}
\begin{tabular}{lcccccccccccccccc}
\toprule
& \multicolumn{4}{c}{\textbf{AMC23}} 
& \multicolumn{4}{c}{\textbf{AIME24}} 
& \multicolumn{4}{c}{\textbf{Math-500}}
& \multicolumn{4}{c}{\textbf{GPQA-D}} \\
\cmidrule(lr){2-5} \cmidrule(lr){6-9} \cmidrule(lr){10-13} \cmidrule(lr){14-17}
\textbf{Method} 
& \textbf{Pass@1} & \textbf{TPS} & \textbf{Latency} & \textbf{Ratio}
& \textbf{Pass@1} & \textbf{TPS} & \textbf{Latency} & \textbf{Ratio}
& \textbf{Pass@1} & \textbf{TPS} & \textbf{Latency} & \textbf{Ratio}
& \textbf{Pass@1} & \textbf{TPS} & \textbf{Latency} & \textbf{Ratio} \\
\midrule

\multicolumn{17}{c}{\textbf{Draft: DeepSeekR1-1.5B \quad Target: DeepSeekR1-32B}} \\
\midrule

\textbf{Random} 
&0.775 &115.90 &49.98  &0.27 
&0.470 &80.94  &124.61 &0.28
&0.840 &109.20 &31.92  &0.20 
&0.308 &65.68  &88.91  &0.31 \\

\textbf{Last-hidden}
&0.865 & \textbf{134.31} &53.69 &0.17 
&0.400 & \textbf{129.65} &\textbf{82.03} &0.14 
&0.874 & \textbf{145.56} &\textbf{32.57} &0.23 
&0.557 &84.61 &69.60 &\textbf{0.21} \\

\textbf{Ours} 
&\textbf{ 0.892} &120.55 &\textbf{ 49.29} &\textbf{0.15}
&\textbf{ 0.553} & 116.27 &\textbf{86.75} &\textbf{0.12}
&\textbf{ 0.910} & 117.33 &\textbf{35.47} &\textbf{0.14}
&\textbf{ 0.573} &\textbf{ 85.28} &\textbf{  66.28}  & 0.22 \\

\midrule
\multicolumn{17}{c}{\textbf{Draft: DeepSeekR1-1.5B \quad Target: QwQ-32B}} \\
\midrule

\textbf{Random} 
&0.758 &54.12 &133.36 &0.31 
&0.460 &53.63 &202.87 &0.29 
&0.814 &58.11 &80.90 &0.28 
&0.286 &52.04 &91.49 &0.21 \\

\textbf{Last-hidden} 
&0.825 &\textbf{146.09} &\textbf{51.87}   &0.31
&0.466 &\textbf{139.84} &\textbf{81.24} &0.29
&0.892 &\textbf{144.79} & \textbf{36.32}  &0.32 
&0.525 &49.68 &156.00 &0.33 \\

\textbf{Ours} 
&\textbf{0.892} & 85.03 & 75.42 &\textbf{0.23}
&\textbf{0.610} & 86.88 & 119.06 & \textbf{0.21}
&\textbf{0.922} & 97.00 & 43.53 &\textbf{0.19}
&\textbf{0.552} &\textbf{73.36} &\textbf{98.68} &\textbf{0.21} \\

\bottomrule
\end{tabular}
}
\caption{
\textbf{Performance comparison} of predictors across four reasoning benchmarks (\textbf{AMC23}, \textbf{AIME24}, \textbf{Math-500}, \textbf{GPQA-D}) in terms of the studied metrics. Each cell reports accuracy (\textbf{Pass@1}), 
average generation speed (\textbf{TPS = Output Tokens / Latency}), 
and inference time (\textbf{Latency}) and we also record the proportion of tokens that were generated by the target model (\textbf{Ratio}).  
}
\label{tab:ab_results}
\end{table*}

\subsection{Models}
To verify both the effectiveness and efficiency of our approach, we conduct experiments with two target models, \textbf{DeepSeek-R1-32B} and \textbf{QwQ-32B}, while using \textbf{DeepSeek-R1-1.5B} as the draft model.



\subsection{Datasets and Metrics} We evaluate our approach on four benchmarks: \textbf{MATH-500}~\cite{hendrycks2021measuring}, \textbf{AIME24}\footnote{\url{https://huggingface.co/datasets/Maxwell-Jia/AIME_2024}}, \textbf{AMC23}\footnote{\url{https://huggingface.co/datasets/zwhe99/amc23}}, and \textbf{GPQA-D}~\cite{rein2024gpqa}. These datasets cover diverse domains of mathematical and scientific reasoning—ranging from competition-level math problems requiring multi-step deduction to expert-level science questions assessing factual and analytical understanding. 

We use three primary metrics: \textbf{Pass@1}, \textbf{Latency}, and \textbf{Token Per Second (TPS)} by following prior studies~\cite{yang2025speculative,pan2025specreason,liao2025reward,yan2024decoding}. The Pass@1 metric measures the proportion of correctly generated answers when generating one candidate solution, reflecting the model’s reasoning and factual reliability. The latency quantifies the total inference latency, including both drafting and verification stages, to evaluate the overall efficiency of speculative decoding. TPS measures the throughput of LLMs and counts the number of output tokens per second (TPS).

\subsection{Baselines}
To evaluate the effectiveness of our approach, we compare our proposed approach with various recent approaches. \textbf{SpecReason} \cite{pan2025specreason} performs speculative reasoning with a lightweight model, using the target model’s verification score to validate and refine the generated outputs.
\textbf{Speculative Thinking} \cite{yang2025speculative} leverages the tendency of reasoning-supportive tokens (e.g., “wait”) to follow structural delimiters and the stronger reflective control of larger models to reduce backtracking and enhance reasoning quality. \textbf{SpecSampling} \cite{leviathan2023fast} is standard token-level Speculative decoding that accelerates the target LLM’s inference without changing its output distribution. Collectively, these baseline methods provide a comprehensive evaluation, covering both segment-level and sentence-level speculative decoding.

\subsection{Training and Implementation Details}
In our experiments, we use the \textbf{SimpleScaling-S1K} dataset \cite{muennighoff2025s1} to train semantic probability predictors, which contains 1,000 high-quality question–answer pairs covering both mathematical and scientific reasoning chains. We use the science-related pairs (i.e., around 400K samples) to train the predictors for GPQA-D, and math-related pairs to train the predictors for math datasets (Math-500, AIME, AMC23). We segment each reasoning chain using the delimiter `\texttt{\textbackslash n\textbackslash n}' and generate 20 responses per example to enhance dataset diversity. Meanwhile we collect the corresponding internal states of LLMs.We eventually collect 180K and 220K samples from the science and math domains, respectively. This offline collection took approximately 8 hours for the target model and 2 hours for the draft model.  For clustering, we employ the \textbf{DeBERTa-Xlarge-MNLI} model to perform iterative pairwise clustering of samples. For training, we adopt a simple three-layer feed-forward neural network (\textbf{FNN}) as the probe for both the target and draft models. Specifically, the target model predictor uses layer dimensions of 5120–2048–512–1, while the draft model predictor uses 1536–512–256–1. We utilize ReLU for all intermediate layers and a Sigmoid activation for the final output layer. The learning rate is set to $1e-3$, and we use the AdamW optimizer \cite{loshchilov2017decoupled} with 
$\beta_1=0.9$ and $\beta_2=0.95$. We use Mean Squared Error (MSE) as our loss function. 
All experiments are conducted on a server with two \textbf{80 GB GPUs} and are implemented using the \textbf{SGLang} framework. Data labeling (clustering via DeBERTa) took roughly 2 hours, while the FNN predictor was trained in just 20 minutes.


\section{Results}\label{results}

\subsection{Comparison with Baselines}\label{sec:rq1}

\textbf{\ourtool (ours) consistently achieves higher pass@1 compared to baselines and competitive efficiency across all datasets and model settings} as shown in Figure~\ref{fig:latency_acc_pairs}. Although SepcThink achieves similar latency as a small model, it also suffers from the lowest pass@1 among all baselines. The reason is that it always accepts the drafted sequences from draft models, while overlooking the refinement from the target model. Compared with SpecReason and SpecSampling, \ourtool consistently achieves higher pass@1 and lower latency across all datasets on both QwQ-32 and DeepSeekR1-32B. Compared with token-level speculative decoding (i.e., SpecSampling), all other sequence-level speculative decoding approaches often achieve lower latency, particularly \ourtool. For instance, \ourtool outperforms SpecSampling by 17.3\% and 7.3\% in terms of pass@1, meanwhile reducing substantial inference time by 40.2\% and 45.3\% on average on DeepSeekR1-32B and QwQ-32B, respectively. However, \ourtool further boosts throughput by 1.67X on the DeepSeekR1-32B and 2.66X on the QwQ-32B compared to SpecSampling. More details can be found in Table~\ref{tab:rasd_results} in the Appendix.

\begin{table*}[ht]
\centering
\resizebox{2\columnwidth}{!}{
\renewcommand{\arraystretch}{1}
\begin{tabular}{lcccccccccccc}
\toprule
& \multicolumn{9}{c}{\textbf{Science $\Rightarrow$ Math}} 
& \multicolumn{3}{c}{\textbf{Math $\Rightarrow$ Science}} \\

\cmidrule(lr){2-10} \cmidrule(lr){11-13}
& \multicolumn{3}{c}{\textbf{AMC23}} 
& \multicolumn{3}{c}{\textbf{AIME24}} 
& \multicolumn{3}{c}{\textbf{Math-500}}
& \multicolumn{3}{c}{\textbf{GPQA-D}} \\

\cmidrule(lr){2-4} \cmidrule(lr){5-7} \cmidrule(lr){8-10} \cmidrule(lr){11-13}
\textbf{Method} 
& \textbf{Pass@1} & \textbf{TPS} & \textbf{Latency}
& \textbf{Pass@1} & \textbf{TPS} & \textbf{Latency}
& \textbf{Pass@1} & \textbf{TPS} & \textbf{Latency}
& \textbf{Pass@1} & \textbf{TPS} & \textbf{Latency} \\
\midrule

\textbf{Within-domain} 
&\textbf{0.892} &120.55 &\textbf{49.29} 
&\textbf{0.553} &116.27 &\textbf{86.75} 
&0.910 &\textbf{117.33} &35.47 
&\textbf{0.573} &\textbf{85.28} &\textbf{66.28} \\

\textbf{Cross-domain} 
&0.875 &\textbf{125.57} &49.90 
&0.533 &\textbf{122.64} &82.38 
&\textbf{0.911} &139.23 &\textbf{31.72} 
&0.552 &70.09 &82.14 \\

\bottomrule
\end{tabular}
}

\caption{
Results of using cross-domain semantic probability predictor. conducted with the \textbf{DeepSeek-R1-1.5B $\rightarrow$ DeepSeek-R1-32B}.}

\label{tab:transfer_results}
\end{table*}

\subsection{Comparison with Original Target and Draft Models.}\label{sec:rq2}

\textbf{Compared with the target model, \ourtool achieves an average speedup of 2.7× and 2.1× on DeepSeekR1-32B and QwQ-32B, respectively, while maintaining comparable pass@1 performance (an average of 4.9\% and 3.9\% pass@1 drops).}
Figure~\ref{fig:latency_acc_pairs} illustrates the effectiveness and efficiency of \ourtool compared with baseline methods across four datasets and two target models. \ourtool consistently accelerates inference, achieving speedups ranging from 2.48× to 3.13× for DeepSeekR1-32B and 1.79× to 2.28× for QwQ-32B. The higher speedup on DeepSeekR1-32B is attributed to using DeepSeekR1-1.5B as the draft model, which belongs to the same model series as the target. Furthermore, \ourtool achieves performance comparable to the target models, with an average performance drop (i.e., pass@1) of 4.9\% and 3.9\% on DeepSeekR1-32B and QwQ-32B, respectively, compared to the target models. However, \ourtool achieves a range of 2.83X and 2.72X higher TPS than targets models on DeepSeekR1-32B and QwQ-32B, respectively.

Compared with the smaller draft model, our approach markedly improves accuracy while shortening reasoning length.
Smaller drafts tend to over-generate repetitive reasoning patterns when they fail to reach a valid intermediate state, producing long, semantically redundant traces that seldom yield the correct answer.
\ourtool guides the draft to produce meaningfully distinct steps aligned with the target model’s reasoning distribution, thereby reducing repetition and enabling the verifier to accept more tokens per step. This yields several-fold accuracy gains over the draft-only baseline (e.g., $+$161.8\% on \textsc{AIME24}) together with shorter reasoning paths.


\subsection{Ablation Analysis.}\label{sec:rq3}
To evaluate the effectiveness of our semantic probability predictor, we conduct an ablation study by comparing it with two variants: (1) \textbf{Random}, where the semantic probabilities for both the target and draft models are assigned random values uniformly sampled from the range [0, 1], while keeping all other components of \ourtool unchanged; and (2) \textbf{Last-hidden}, where the predictor utilizes only the hidden states from the final layer rather than aggregating representations from all layers.

Table~\ref{tab:ab_results} presents the comparison results. Compared to Random, our predictor achieves substantial improvements across all evaluation metrics on all models, demonstrating its effectiveness in accurately routing inference between the target and draft models. Compared with Last-hidden, our predictor consistently yields higher Pass@1 scores across all datasets and models—for example, improving Pass@1 from 0.400 to 0.553 and from 0.466 to 0.610 on AIME24 for DeepSeekR1-32B and QwQ-32B, respectively. 
    However, Last-hidden has the highest throughput (although it use the target model to generate more tokens), as it only accesses the final layer’s hidden states, whereas our predictor incurs additional computational overhead by extracting features from all layers. This observation highlights a \textbf{trade-off between effectiveness and efficiency: incorporating multi-layer information enhances predictive accuracy but reduces throughput.} Practitioners can therefore choose between the two designs based on their performance requirements and latency constraints in practical deployment scenarios.

\subsection{Generalization of Semantic Probability
Predictor}\label{sec:rq4}

To evaluate the generalization ability of the proposed semantic probability predictor, we conduct cross-domain prediction experiments between the math and science domains.
Specifically, we use the predictor trained on science data (i.e., the science subset of SimpleScaling-S1K) to predict the probabilities for math tasks (AMC23, AIME24, and Math-500), and the predictor trained on math data to predict the probabilities for science tasks (GPQA-D).

Table~\ref{tab:transfer_results} reports the results for both within-domain (predictor and tasks from the same domain) and cross-domain settings on DeepSeek-R1-32B.
Compared to within-domain prediction, cross-domain performance shows a slight decrease on AMC23 (from 0.892 to 0.875), AIME24 (from 0.553 to 0.533), and GPQA-D (from 0.573 to 0.552), while remaining comparable on Math-500 (from 0.910 to 0.911). \textbf{These results suggest that the semantic probability predictor based on internal states of models generalizes well across domains}.



\section{Discussion}\label{sec:dis}

\subsection{Overhead}

Across all tasks and model pairs, the semantic probability prediction adds only 0.12 - 0.22 seconds to the decoding process, less than 0.4\% of the inference time. In terms of memory usage, the target model and draft model occupy 73.7 GB and 26.3 GB, respectively, while the two semantic probability predictors together require only 0.58 GB. Therefore, the predictors contribute less than 1\% additional memory overhead relative to the models themselves.

\subsection{Robustness to semantic clustering noise}

To evaluate the system's robustness against potential errors in semantic clustering (e.g., DeBERTa-based biases), a noise injection stress test was conducted on AMC23 using DeepSeekR1-1.5B (Draft) and DeepSeekR1-32B (Target). Cluster assignments were manually corrupted by randomly reassigning $x\%$ of samples to incorrect clusters. The predictor remains highly effective under 10\% noise, showing only a marginal decrease in Pass@1 (from 0.892 to 0.866). Performance significantly degrades at 30\% noise, falling below the ``Random Assignment'' baseline (0.775). This indicates that while the model is resilient to minor semantic inaccuracies, the predictor relies on a minimum threshold of cluster purity to outperform a naive random baseline. Interestingly, TPS remains high even as accuracy drops, suggesting the predictor still attempts to generate drafts, though their quality (Pass@1) suffers.

\begin{table}[h!]
\centering
\caption{Noise Injection Stress Test on AMC23}
\label{tab:noise_injection}
\begin{tabular}{cccc}
\hline
\textbf{Noise} & \textbf{Pass@1} & \textbf{Latency (s)} & \textbf{TPS} \\ 
\hline
0\%            & 0.892           & 49.30                & 120.60       \\
10\%           & 0.866           & 56.86                & 145.39       \\
30\%           & 0.717           & 58.07                & 155.50       \\
50\%           & 0.700           & 61.61                & 148.25       \\
\hline
\end{tabular}
\end{table}

\section{Conclusion}\label{sec:conclusion}
This work introduces a semantic-aware speculative decoding that advances beyond token-level by capturing meaning rather than surface token patterns. Through a novel semantic probability estimation mechanism based on internal model representations, \ourtool effectively identifies semantically valid sequences during verification. Our extensive experiments demonstrate that it substantially accelerates inference while preserving semantic fidelity, and achieves up to 2.7× speedup over standard decoding. These results highlight the potential of semantic-aware speculative decoding as a new direction for efficient LLM inference.

\section{Limitations}
In this paper, we evaluated our semantic-level speculative decoding on two pairs of open source models, and four widely used benchmarks. Our findings might be not generalized to other models, we encourage future research to measure the effective of our approach on more models. 







\bibliography{custom}

\newpage
\appendix

\section{Appendix}
\label{sec:appendix}
\subsection{Empirical evidence of semantically equivalent but lexically different sequence rejection}\label{sec:falserejection}

To quantify the frequency of false rejections, where sequences are semantically equivalent but lexically divergent, we conducted a controlled analysis using 100 samples from SimpleScaling-S1K dataset, where we collected training data.

We segmented reasoning traces into discrete steps using \texttt{\textbackslash n\textbackslash n} as the delimiter. For each prefix (original query + multiple reasoning steps), we generated the subsequent reasoning step under two conditions:

\begin{itemize}
    \item \textbf{Strict Lexical Verification:} Traditional speculative decoding (rejection on any token mismatch).
    \item \textbf{Unconstrained Draft Generation: }The draft model generates the sequence without any target model intervention or rejection 
\end{itemize}

We collected 20,496 pairs of generated outputs for comparison. We then evaluated the semantic equivalence (using the bi-directional entailment algorithm described in Section~\ref{sec:method}) of these pairs to identify ``false rejections''.

45.8\% (9,396/20,496) of the output pairs have the same semantic meaning, and among them, 56.5\% (5,315/9,396) pairs have the same semantic meaning but lexical mismatches, where at least one token was rejected in Strict Lexical Verification. In other words, in 56.5\% of cases, the draft and target models agree on the reasoning intent, while traditional speculative decoding triggered a false rejection due to minor token-level differences. Crucially, in these false rejection instances, we observed an average of 3.25 tokens rejected per sequence. \textbf{This high frequency of false rejections directly translates to a latency bottleneck, as the target model is forced to re-generate sequences that were already semantically correct. This validates the necessity of SemanticSpec in bypassing strict lexical constraints to unlock higher effective throughput.}

\subsection{Impact of drafted sequence length $\gamma$}

\begin{figure}[ht]
    \centering

    \includegraphics[width=1.01\linewidth]{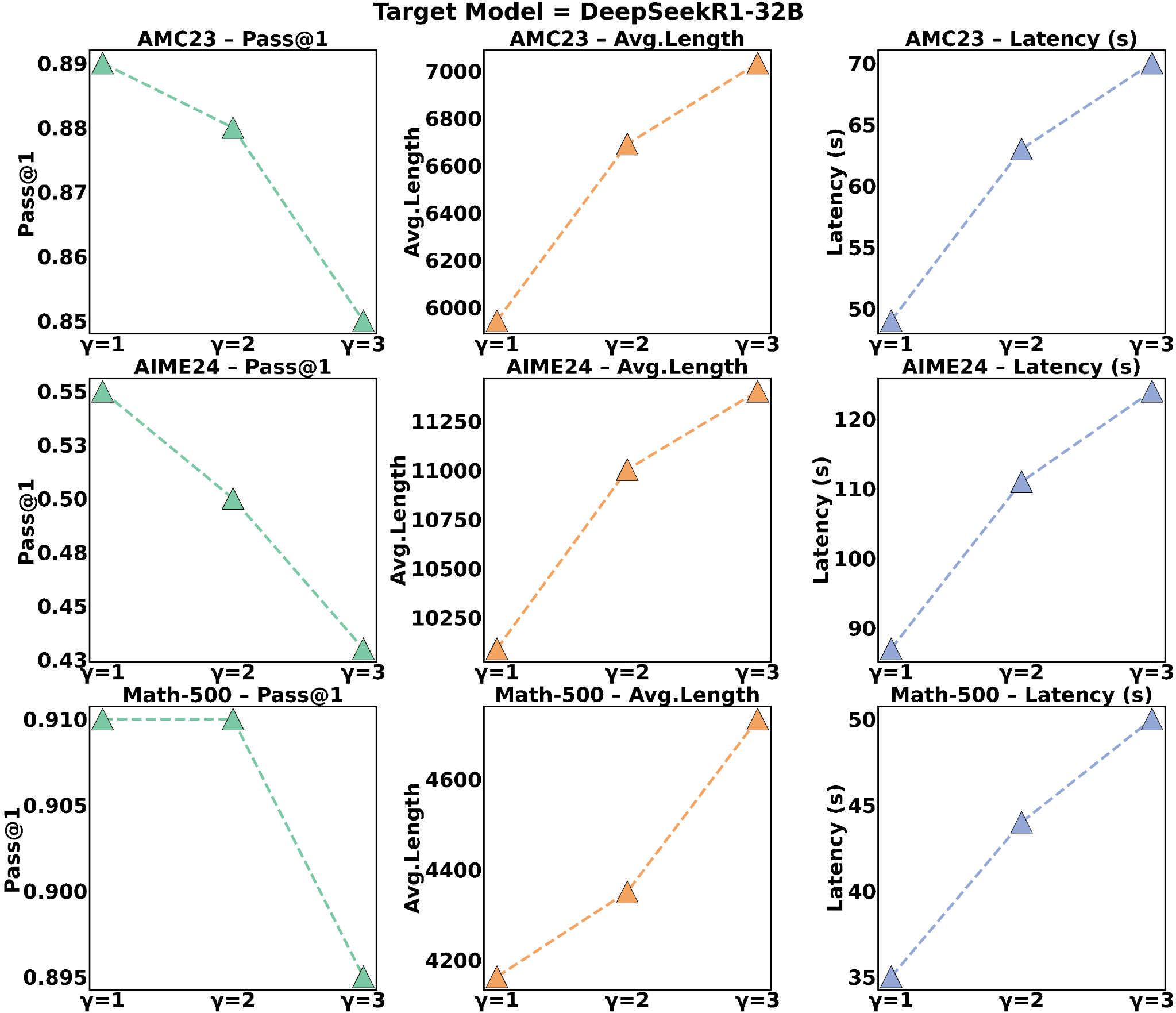} 
    \caption{
    Performance comparison under different size of drafted sequence ($\gamma = {1,2,3}$). 
    }
    \label{fig:impactOfgamma}
\end{figure}

To further investigate the impact of drafted sequence length (i.e., impact of hyper-parameter $\gamma$), we evaluate the performance of our semantic-level speculative decoding under different values ($\gamma \in {1, 2, 3}$) across all datasets. As observed in Figure~\ref{fig:impactOfgamma}, as the length of the drafted sequence increases, both the effectiveness and efficiency of our semantic-level speculative decoding consistently decreases across all datasets, which suggests to use small $\gamma$ in practice. This is reasonable, since we use   `\texttt{\textbackslash n\textbackslash n}' to split sequences, a large $\gamma$ usually results in a long sequence which probably expresses various semantic meanings, which reduce the accepted rate of the drafted sequences.

\subsection{Impact of sequence splitting strategy}

\begin{figure}                                                                                                                                                                 
    \centering
    \includegraphics[width=0.48\textwidth]{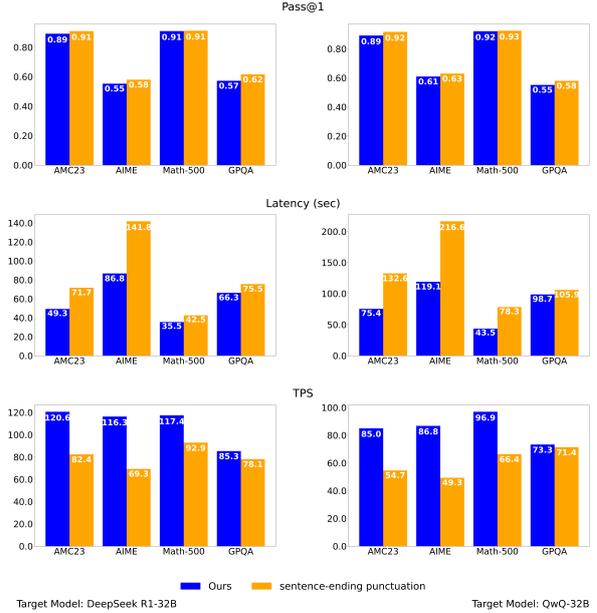}

    \caption{The impact of different splitting strategies ours vs. sentence-ending punctuation.}   \label{fig:Lambda_impact}

\end{figure}

In default setting, we use \textbf{`\textbackslash n\textbackslash n'} as the split to separate sequence, which probably lead to relatively coarse-granularity. To investigate the impact of smaller granularity, we also investigate to use \textbf{sentence-ending punctuation} (such as \textbf{.?!}) as the split. As shown in Figure\ref{fig:Lambda_impact}, increasing the granularity consistently improves pass@1 across all datasets slightly, while TPS decreases significantly. Finer-grained split triggers more frequent validations, resulting in a more constrained generation process. 

\subsection{Generalization to other platforms}

To demonstrate the generalization ability of our method, we also applied it within the Hugging Face framework. Table \ref{tab:absults} shows that our method reduces output length while maintaining performance competitive with the target model, indicating that it can be applied across a variety of frameworks as long as the hidden states are accessible.



\begin{table}[ht]
\centering
\resizebox{1\columnwidth}{!}{
\renewcommand{\arraystretch}{1}
\begin{tabular}{lcccccc}
\toprule
& \multicolumn{3}{c}{\textbf{AIME24}} 
& \multicolumn{3}{c}{\textbf{Math-500}} \\
\cmidrule(lr){2-4} \cmidrule(lr){5-7}
\textbf{Method} 
& \textbf{Pass@1} & \textbf{TPS} & \textbf{Latency}
& \textbf{Pass@1} & \textbf{TPS} & \textbf{Latency} \\
\midrule

\textbf{Draft Model} 
&0.211 &21.99 &519.35  
&0.738  &40.49 &127.73  \\

\textbf{Target Model} 
&0.550 &11.86 &735.75
&0.945 &11.09 &333.01 \\

\textbf{Ours} 
&0.520 &14.80 &653.05
&0.935 &20.51 &243.92 \\

\bottomrule
\end{tabular}
}
\caption{The performance across two reasoning benchmarks 
(\textbf{AIME24}, \textbf{Math-500}). 
Each cell reports accuracy (\textbf{Pass@1}), 
updated throughput (\textbf{TPS / Latency} on Hugging Face.), 
and inference time (\textbf{Latency}). 
}
\label{tab:absults}
\end{table}

\subsection{Detailed results of the studied speculative decoding methods}
Table~\ref{tab:rasd_results} presents the detailed performance of the studied speculative decoding approaches in terms of Pass@1, TPS, and Latency.
\vspace{-1mm} 
\begin{table*}[t!]
\centering
\resizebox{2.05\columnwidth}{!}{
\Huge
\setlength{\tabcolsep}{4pt}
\renewcommand{\arraystretch}{1.2}
\begin{tabular}{llcccccccccccc}

\toprule
 & \multicolumn{3}{c}{\textbf{AMC23}} 
   & \multicolumn{3}{c}{\textbf{AIME24}} 
   & \multicolumn{3}{c}{\textbf{Math-500}}
   & \multicolumn{3}{c}{\textbf{GPQA-D}} \\
\cmidrule(lr){2-4} \cmidrule(lr){5-7} \cmidrule(lr){8-10} \cmidrule(lr){11-13}
\textbf{Method} 
& \textbf{Pass@1} & \textbf{TPS} & \textbf{Latency} 
& \textbf{Pass@1} & \textbf{TPS} & \textbf{Latency} 
& \textbf{Pass@1} & \textbf{TPS} & \textbf{Latency} 
& \textbf{Pass@1} & \textbf{TPS} & \textbf{Latency} \\
\midrule

\multicolumn{13}{l}{\hspace{15em}\textbf{Draft: DeepseekR1-1.5B \quad Target: DeepseekR1-32B}} \\
\midrule
 \textbf{Draft Model} &0.658 &378.30 &19.59 &0.233 &380.67 &29.45 &0.8142 &380.51 &12.70 &0.338 &371.93 & 21.30 \\ 
 \textbf{Target Model} &0.925 &34.78 &154.34 & 0.621 &41.03 & 214.76 &0.928 &39.21 & 89.91 & 0.593 &40.65 & 170.87 \\
\textbf{SpecReason} & 0.775 & 125.92 & 49.98 & 0.470 & 80.95 & 124.61 & 0.840 & 149.19 & 31.92 & 0.281 & 52.79 & 159.33 \\
\textbf{SpecThinking} & 0.650 & \textbf{284.59} & \textbf{24.99} & 0.241 & \textbf{272.79} & \textbf{37.09} & 0.823 & \textbf{284.34} & \textbf{15.94} & 0.315 & \textbf{250.31} & \textbf{44.88} \\
\textbf{SpecSampling} & 0.842 & 65.73 & 91.40 & 0.443 & 53.29 & 191.45 & 0.880 & 68.58 & 59.32 & 0.424 & 73.00 & 82.78 \\
\textbf{Ours} & \textbf{0.892 } & 120.58 & 49.29 & 
\textbf{0.553  } & 116.32 & 86.75 &
\textbf{0.910  } & 117.36 & 35.47 &
\textbf{0.5725  } & 85.30 & 66.28 \\
\midrule

\multicolumn{13}{l}{\hspace{15em}\textbf{Draft: DeepseekR1-1.5B \quad Target: QwQ-32B}} \\
\midrule
\textbf{Draft Model} &0.658 &378.30 &19.59 &0.233 &380.67 &29.45 &0.8142 &380.51 &12.70 &0.338 &371.93 & 21.30\\ 
 \textbf{Target Model} &0.925 & 40.62 & 165.02 & 0.63 & 40.62 & 250.75 & 0.935 & 40.71 & 99.04 & 0.5959 & 39.73 & 176.25  \\
\textbf{SpecReason} & 0.857 & 64.52 & 98.59 & 0.610 & 45.96 & 205.70 & 0.770 & 76.02 & 53.66 & 0.310 & 68.83 & 109.64 \\
\textbf{SpecThinking} & 0.614 & 309.55 & 30.20 & 0.204 & 285.20 & 45.87 & 0.710 & 312.77 & 27.99 & 0.310 & 250.31 & 44.88 \\
\textbf{SpecSampling} & 0.825 & 58.53 & 120.39 & 0.540 & 34.25 & 316.19 & 0.895 & 52.61 & 90.33 & 0.5250 & 37.59 & 194.01 \\
\textbf{Ours} & \textbf{0.892} & \textbf{85.00} & \textbf{75.42} & \textbf{0.610} & \textbf{86.79} & \textbf{119.06} & \textbf{0.9217} & \textbf{96.95} & \textbf{43.53} & \textbf{0.552} & \textbf{73.32} & \textbf{98.68} \\
\midrule

\end{tabular}
}
\caption{Performance comparison of different methods across \textbf{Math-500}, \textbf{AMC23}, \textbf{AIME24}, and \textbf{GPQA-D}. The the cells achieving the best result among ours and baselines are marked as bold.}
\label{tab:rasd_results}
\end{table*}








\onecolumn
\newpage
\thispagestyle{plain}
\setlength{\topmargin}{-1.5cm}
\setlength{\textheight}{25cm}
\setlength{\parskip}{4pt}
\setlength{\parindent}{1em}

\begin{center}
\textbf{Appendix: Example from AMC23 Dataset}
\end{center}
\vspace*{-7cm}


\begin{tcolorbox}[
enhanced,
breakable,
colback=white!5,
colframe=gray!60!white,
fonttitle=\bfseries,
coltitle=black,
left=6pt, right=6pt, top=5pt, bottom=5pt,
width=\textwidth,
title=\textbf{Question},
before skip=0pt, after skip=8pt,
arc=0pt, boxrule=0.7pt
]

Flora the frog starts at 0 on the number line and makes a sequence of jumps to the right. 
In any one jump, independent of previous jumps, Flora leaps a positive integer distance $m$ 
with probability $\frac{1}{2^m}$. 
What is the probability that Flora will eventually land at 10? 
Write the answer as a simplified fraction $\frac{m}{n}$, find $m+n$.
\end{tcolorbox}
\begin{tcolorbox}[
enhanced,
breakable,
colback=red!5,
colframe=red!60!white,
fonttitle=\bfseries,
coltitle=black,
left=6pt, right=6pt, top=5pt, bottom=5pt,
width=\textwidth,
title=\textbf{Speculative Thinking(wrong!)},
before skip=0pt, after skip=8pt,
arc=0pt, boxrule=0.7pt
]
Alright, so I need to find the probability that Flora the frog will eventually land on 10 starting from 0. 
She can jump any positive integer distance $m$ with probability $\frac{1}{2^m}$ each time, and each jump is independent. 

First, note that once she jumps past 10, she can never return, since all jumps are to the right. 
Thus, the only way Flora can ever land on 10 is if the sum of her jumps equals exactly 10 at some point.

Let $P(n)$ be the probability that starting from position $n$, the frog will eventually reach 10. 
We have:
\[
P(n) = \sum_{m=1}^\infty \frac{1}{2^m} P(n+m),
\]
with $P(10)=1$ and $P(n)=0$ for $n>10$. \\

We can recursively compute backward:
\[
P(9)=\tfrac{1}{2}, \; P(8)=\tfrac{1}{2}, \; \dots, \; P(1)=\tfrac{1}{2}.
\]
Then,
\[
P(0) = \sum_{m=1}^{\infty} \frac{1}{2^m} P(m) 
     = \frac{1}{2} \sum_{m=1}^{9} \frac{1}{2^m}.
\]
The finite geometric series gives:
\[
\sum_{m=1}^{9} \frac{1}{2^m} = \frac{1/2(1 - (1/2)^9)}{1 - 1/2} = \frac{511}{512}.
\]
Hence,
\[
P(0) = \frac{1}{2} \times \frac{511}{512} = \frac{511}{1024}.
\]

Therefore, the probability that Flora will eventually land on 10 is $\frac{511}{1024}$, 
so $m = 511$, $n = 1024$, and $m+n = 1535$.

\textit{\textcolor{gray}{</think>}} \\[0.4em]

\textbf{Final Answer:}
\[
\boxed{\frac{511}{1024}}, \quad \boxed{m+n = 1535}.
\]

\end{tcolorbox}

\begin{tcolorbox}[
enhanced,
breakable,
colframe=yellow!40!black,
fonttitle=\bfseries,
title=\textbf{Speculative Sampling(9162 tokens, 102s)},
left=6pt, right=6pt, top=4pt, bottom=4pt,
width=\textwidth,
before skip=2pt, after skip=8pt,
arc=0pt, boxrule=0.6pt
]
Okay, so I have this problem where Flora the frog starts at 0 on the number line and jumps to the right. Each jump she makes is a positive integer distance \( m \), and the probability of jumping \( \frac{1}{2^m} \). I need to find the probability that she will eventually land on 10. The answer should be written as a simplified fraction \( \frac{m}{n} \), and then I need to find \( m + n \).

Hmm\textcolor{red}{(Okay)}, let me think about how to approach \textcolor{red}{(solve)} this. It seems like a probability problem involving \textcolor{red}{(about)} jumps on a number line. Since each jump is independent\textcolor{red}{(uncorrelated)}, the probability of reaching 10 depends\textcolor{red}{(relies)} on the probabilities of all possible sequences of jumps that sum \textcolor{red}{(total)} up to 10. So\textcolor{red}{(Thus)}, maybe\textcolor{red}{(I)} I can model \textcolor{red}{(take)}this as a recursive probability problem.

Let me denote \( P(k) \) as the probability that Flora will eventually  \textcolor{red}{(land)} land on position \( k \). So,  \textcolor{red}{(Thus)} I need \textcolor{red}{(should)} to find \( P(10) \).

Since \textcolor{red}{(Because)} Flora can \textcolor{red}{(could)} jump any positive integer distance \( m \) with probability \( \frac{1}{2^m} \), the probability of landing on position \( k \) can \textcolor{red}{(could)} be thought of as the sum of the probabilities of all\textcolor{red}{(whole)} possible ways to reach \textcolor{red}{(find)} \( k \) in one jump from some \textcolor{red}{(any)}  position \( j \), where \( j \) is less than \( k \).

Wait\textcolor{red}{(Hold)}, actually\textcolor{red}{(that's)}, that's\textcolor{red}{(likely)} more like the probability of reaching \( k \) for the first time. But \textcolor{red}{(However)} the problem\textcolor{red}{(question)} says "eventually land at 10," which might \textcolor{red}{(include)} include\textcolor{red}{(have)} landing on 10 multiple times. But \textcolor{red}{(Wait)} I think \textcolor{red}{(can)} in this context, it's just about whether\textcolor{red}{(if)} she ever lands on 10, not \textcolor{red}{(but)} necessarily the first time. So, perhaps\textcolor{red}{(maybe)} it's the probability that \textcolor{red}{(she)} she reaches 10 in some finite number of jumps.

[...] 
Therefore, starting from 0, \( P(0) = \frac{1}{2} \). Therefore, the probability is \( \frac{1}{2} \), so \( m = 1 \), \( n = 2 \), and \( m + n = 3 \).

I think this is correct. Each position before 10 has a 50

**Final Answer**
The probability is \(\boxed{3}\).
\end{tcolorbox}

\begin{tcolorbox}[
enhanced,
breakable,
colback=green!5,
colframe=green!40!black,
fonttitle=\bfseries,
title=\textbf{\ourtool(6092 tokens, 64s)},
left=6pt, right=6pt, top=4pt, bottom=4pt,
width=\textwidth,
before skip=2pt, after skip=4pt,
arc=0pt, boxrule=0.6pt
]
Okay, so I have this problem about Flora the frog who starts at 0 on the number line and makes jumps only to the right. Each jump she makes is a positive integer distance $m$ with probability $1/(2^m)$. I need to find the probability that she will eventually land exactly on 10. Hmm, that sounds like it involves some probability theory and maybe recursion or generating functions.

[...] 
\textcolor{red}{But notice that the inner sum is $\sum_{k=0}^{\infty} P(k)\, 2^k\, x^{n - k}$. 
Wait, because for each $k$, it's multiplied by $x^n$ and $2^k$,}\textcolor{red}{so perhaps we can factor $x^n$ as $x^{n - k} \cdot x^k$. Hmm, maybe not.
}

\textcolor{green}{Wait, perhaps we can swap the order of summation on the right-hand side. Let's try:}
[...] 
\textcolor{red}{Wait, but let me think again. If the frog can jump any positive integer distance, then the probability of eventually reaching 10 is 1/2? That seems high, but given that the frog can potentially jump any distance, the probability might be higher.}
\textcolor{green}{Wait, let's think about the recurrence relation. For $n \ge 1$, 
$P(n) = \sum_{m=1}^{n} P(n - m) \cdot \frac{1}{2^m}$. 
So, if we assume that $P(k) = \frac{1}{2}$ for all $k \ge 1$, then:
}
[...] 
\textcolor{red}{In such a case, the probability of reaching state 10 is the same for all states n >=1, due to the symmetry in the jump probabilities. Wait, is that true?"}
\textcolor{green}{In such a chain, since all jumps are to the right, and the frog can potentially jump over 10 in a single jump, the question is whether the frog will ever land exactly on 10.}
[...] 
\textcolor{red}{Wait, but that seems a bit counterintuitive because, in reality, the frog has to make a finite number of jumps to reach 10, but the probability of making such jumps is decreasing exponentially.}
\textcolor{green}{Alternatively, another way to think about it is that for each position $n \ge 1$, the probability of ever reaching $n$ is $1/2$. That seems to be the case here because of the generating function.} 

So, perhaps the answer is 1/2 for P(10). So, $m=1$, $n=2$, $m + n=3$.

\[
\boxed{3}
\]
</think>

Flora the frog starts at 0 on the number line and makes jumps to the right. Each jump distance \( m \) has a probability of \( \frac{1}{2^m} \). We need to find the probability that Flora will eventually land on 10.

We denote \( P(n) \) as the probability that Flora will eventually land on position \( n \). We need to find \( P(10) \).

The recurrence relation for \( P(n) \) is:
\[
P(n) = \sum_{m=1}^n \frac{P(n - m)}{2^m}
\]
with the base case \( P(0) = 1 \).

We derived the generating function \( G(x) \) for the sequence \( P(n) \) and found that:
\[
G(x) = \frac{2 - x}{2(1 - x)}
\]

Expanding \( G(x) \) as a power series, we found that:
\[
G(x) = 1 + \frac{1}{2}x + \frac{1}{2}x^2 + \frac{1}{2}x^3 + \cdots
\]

This expansion shows that for all \( n \geq 1 \), the coefficient of \( x^n \) is \( \frac{1}{2} \). Therefore, the probability \( P(n) \) for any \( n \geq 1 \) is \( \frac{1}{2} \).

Thus, the probability that Flora will eventually land on 10 is \( \frac{1}{2} \). The fraction \( \frac{1}{2} \) is already in its simplest form, so \( m = 1 \) and \( n = 2 \). Therefore, \( m + n = 1 + 2 = 3 \).

The final answer is:
\[
\boxed{3}
\]
\end{tcolorbox}

\vspace{0.5em}
\noindent An example from the AMC23 dataset showing how \ourtool, 
\textbf{speculative sampling}, and \textbf{speculative thinking} 
produce different responses to the same input. In \textbf{speculative sampling}, the red word denotes the token replaced by its left neighbor. In contrast, \textbf{\ourtool}  replaces entire sentences (red part) with the target model’s output (green part) when the model judges them to be untruthful. From the example, we can see that \textbf{speculative sampling} is very restricted. However, \textbf{\ourtool} implicitly identifies abnormal sequences and replaces them more appropriately.

\newpage
\twocolumn

\end{document}